%% file: main.tex
\documentclass{article}
\usepackage[preprint]{spconf}
\usepackage{amsmath,graphicx,hyperref}
\usepackage{multirow, multicol, makecell}

\copyrightnotice{\copyright~2026 IEEE}
\toappear{To appear in {\it Proceedings of IEEE ICASSP 2026, Barcelona, Spain, May 3--8, 2026}}

\title{Targeted Fine-Tuning of DNN-Based Receivers via Influence Functions}
\twoauthors
  {Marko Tuononen\sthanks{Author would like to thank Dani Korpi and Vesa Starck.}, Heikki Penttinen\sthanks{Author would like to thank Esa Ollila.}}
	{Nokia Networks\\
    Nokia Group\\
    Espoo, Finland}
  {Ville Hautamäki\sthanks{Author was partially supported by Jane and Aatos Erkko Foundation.}}
	{School of Computing\\
    University of Eastern Finland\\
    Joensuu, Finland}
\begin{document}
\ninept

\maketitle

\begin{abstract}
We present the first use of influence functions for deep learning-based wireless receivers. Applied to DeepRx, a fully convolutional receiver, influence analysis reveals which training samples drive bit predictions, enabling targeted fine-tuning of poorly performing cases. We show that $\ell$-relative influence with capacity-like binary cross-entropy loss and first-order updates on beneficial samples most consistently improves bit error rate toward genie-aided performance, outperforming random fine-tuning in single-target scenarios. Multi-target adaptation proved less effective, underscoring open challenges. Beyond experiments, we connect influence to self-influence corrections and propose a second-order, influence-aligned update strategy. Our results establish influence functions as both an interpretability tool and a basis for efficient receiver adaptation.
\end{abstract}


\section{Introduction}
\label{sec:introduction}
Neural networks are increasingly deployed across science, technology, and society~\cite{allen2024}, often in high-stakes environments~\cite{jumper2021alphafold}. Despite their success, their internal decision-making processes remain opaque: a single prediction may involve billions of operations, making human interpretation infeasible~\cite{molnar2022}. This opacity complicates debugging, trust, and accountability~\cite{huyen2022,adebayo2020debugging}, especially where model behavior must align with physical or safety constraints. As a result, interpretable machine learning has become both a technical necessity and a societal requirement~\cite{tuononen2025nap}, emphasized by regulatory frameworks such as the EU AI Act~\cite{euaiact2024} and AI Ethics Guidelines~\cite{euaiethics2019}. While interpretability methods were largely developed in Computer Vision (CV) and Natural Language Processing (NLP), they are increasingly applied to data-rich scientific and engineering domains where transparency is essential for discovery and reliable deployment~\cite{allen2024}.

Wireless communication is one such domain. As 5G matures and 6G research accelerates~\cite{farhadi2025}, physical-layer processing faces unprecedented demands for efficiency, reliability, and adaptability~\cite{nokia2024envisioning6g}. Classical pipelines~\cite{goldsmith2005,tse2005}--built around explicit processing blocks--are approaching their theoretical and practical limits, motivating a shift toward data-driven receivers~\cite{hoydis2021toward}. Architectures like DeepRx~\cite{honkala2021deeprx} replace these blocks with a fully convolutional network that processes raw frequency-domain signals to decode transmitted bits under dynamic channel conditions. While effective, such models introduce a new challenge: interpretability~\cite{tuononen2025nucpi,belgiovine2025atlas}. Physical-layer inputs are non-semantic, combining transmitted data with environmental effects (noise, fading, interference), rendering most standard interpretability methods ineffective~\cite{chergui2025,fiandrino2022}.

\begin{figure}[t]
\centering
\includegraphics[width=0.97\columnwidth]{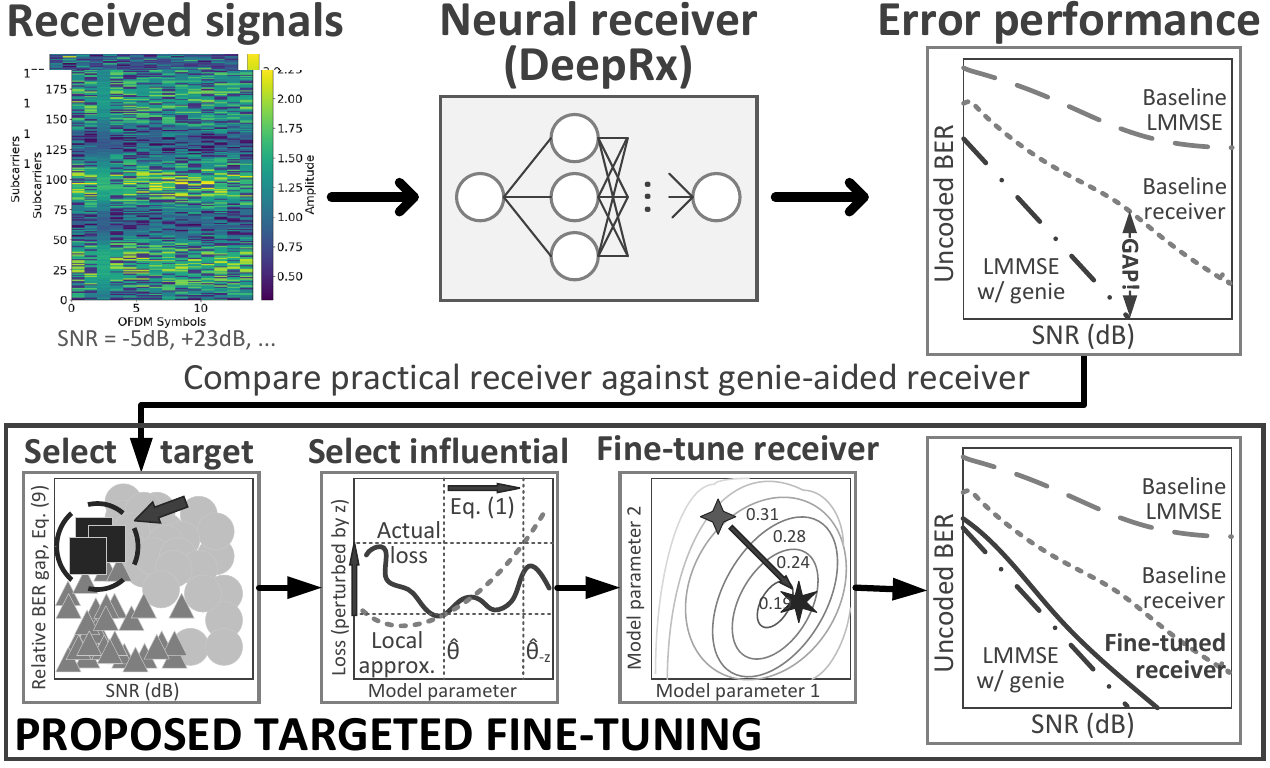}
\caption{Illustration of the proposed approach: (top) neural receiver processes received signals to estimate transmitted bits with baseline performance; (bottom) targeted fine-tuning via influence functions leverages the most beneficial training instances, improving receiver performance and narrowing the gap to the genie-aided benchmark.}
\label{fig:high_level}
\end{figure}

Since machine learning reliability hinges on data~\cite{reddi2021dataengineering}, understanding how individual samples influence predictions is critical: harmful or unrepresentative training points may silently degrade performance~\cite{hammoudeh2024}. \emph{Influence Functions (IFs)}~\cite{koh2017if}--long used in statistics~\cite{hampel1974influence} and later adapted to deep learning~\cite{koh2017if}--estimate how perturbing a training instance shifts model predictions. Beyond interpretability, they provide a principled way to trace prediction errors back to their causes and guide targeted model adaptation~\cite{schioppa2023whatifdo}. While IFs are widely studied in CV and NLP~\cite{koh2017if,han2020}, they remain unexplored in neural wireless receivers.

In this paper we apply influence functions~\cite{koh2017if} to deep-learning based wireless receivers for the first time. Applied to DeepRx~\cite{honkala2021deeprx}, a fully convolutional receiver, we build on recent advances: scalable approximations for large models~\cite{schioppa2022scalingif}, cross-loss formulations aligned with task metrics~\cite{silva2022crosslossif}, local variants that highlight relative influence~\cite{barshan2020relatif}, and corrections that remove self-influence~\cite{zou2025newfluence}. These tools allow us to analyze how training data affect bit predictions and to introduce targeted fine-tuning that adapts receivers using their most beneficial training samples (Fig.~\ref{fig:high_level}, Sec.~\ref{sec:proposed_fine_tuning}). We further propose methodological refinements to influence estimation and fine-tuning, and evaluate both single- and multi-target adaptation. Overall, our results establish influence functions as both a diagnostic tool and a basis for efficient receiver adaptation, improving performance toward genie-aided benchmarks while opening new directions for adaptive and interpretable wireless systems.

\section{Background}
\label{sec:background}
Influence Functions (IFs) originate from robust statistics~\cite{hampel1974influence} and were adapted to deep learning by Koh and Liang~\cite{koh2017if}. They estimate how perturbing a training instance $z_i$, $i \in\{1,\ldots,n\}$, changes the learned parameters and, in turn, the loss on a test point $z_{\text{test}}$. For empirical risk minimization with loss $\ell$, let $\hat{\theta}$ denote the learned parameters. The parameter change from upweighting $z_i$ by an infinitesimal amount $\epsilon$ is~\cite[Eq.1]{koh2017if}
\begin{equation}
    \frac{d\hat{\theta}_{\epsilon,z_i}}{d\epsilon}\Big|_{\epsilon=0}
    = - H_{\hat{\theta}}^{-1} \nabla_\theta \ell(z_i,\hat{\theta}),
\end{equation}
where $H_{\hat{\theta}} = \frac{1}{n}\sum_{i=1}^n \nabla^2_\theta \ell(z_i,\hat{\theta})$ is the Hessian of the empirical loss. The influence of $z_i$ on the test loss is then~\cite[Eq.2]{koh2017if}
\begin{equation}\label{eq:classic_if}
    \mathcal{I}_{\text{Classic}}(z_{\text{test}}, z_i)
    = - \nabla_\theta \ell(z_{\text{test}},\hat{\theta})^\top 
      H_{\hat{\theta}}^{-1} \nabla_\theta \ell(z_i,\hat{\theta}).
\end{equation}
Samples with negative influence reduce test loss (beneficial), while positive influence samples increase it (harmful).

\subsection{Cross-Loss Influence -- Aligning with Task Metrics}
Cross-loss IF (CLIF)~\cite{silva2022crosslossif} generalizes the framework by decoupling the training and evaluation losses. For training loss $\ell_{\text{train}}$ and evaluation loss $\ell_{\text{eval}}$, the influence is~\cite[Theorem 1]{silva2022crosslossif}
\begin{equation}\label{eq:cross_loss_if}
    \mathcal{I}_{\text{CLIF}}(z_{\text{test}}, z_i)
    = - \nabla_\theta \ell_{\text{eval}}(z_{\text{test}},\hat{\theta})^\top 
      H_{\hat{\theta}}^{-1} \nabla_\theta \ell_{\text{train}}(z_i,\hat{\theta}).
\end{equation}
This is particularly useful when the system-level metric (e.g., BER) differs from the training objective (e.g., binary cross-entropy).

\subsection{Relative Influence -- Highlighting Local Influence}
Relative influence functions constrain the comparison either in parameter space or in loss space~\cite[Definitions 3.2 and 3.4]{barshan2020relatif}:
\begin{align}
    \mathcal{I}_{\theta\text{-relative}}(z_{\text{test}}, z_i) 
    &= \frac{\mathcal{I}_{\text{Classic}}(z_{\text{test}}, z_i)}{\|H_{\hat{\theta}}^{-1} \nabla_\theta \ell(z_i,\hat{\theta})\|}, \label{eq:theta_relative_if} \\
    \mathcal{I}_{\ell\text{-relative}}(z_{\text{test}}, z_i) 
    &= \frac{\mathcal{I}_{\text{Classic}}(z_{\text{test}}, z_i)}{\sqrt{\mathcal{I}_{\text{Classic}}(z_i, z_i)}} \label{eq:ell_relative_if}.
\end{align}
These normalize influence, providing a localized view of it and making comparisons across models or datasets more interpretable.

\subsection{Newfluence -- Removing Self-Influence}
Building on Newfluence~\cite{zou2025newfluence} we correct influence estimates using the self-influence $\mathcal{I}_{\text{Classic}}(z_i, z_i)$ of $z_i$. The corrected influence is
\begin{equation}\label{eq:newfluence_if}
    \mathcal{I}_{\text{Newfluence}}(z_{\text{test}}, z_i)
    = \frac{\mathcal{I}_{\text{Classic}}(z_{\text{test}}, z_i)}{1 + \mathcal{I}_{\text{Classic}}(z_i, z_i)}.
\end{equation}
This correction--related to the Sherman–Morrison update~\cite{sherman1950adjustment} and regression diagnostics~\cite{cook1982residuals}--mitigates high-leverage points and, in theory, yields more accurate estimates without sacrificing efficiency.

\subsection{Influence Approximations -- Scaling to Deep Models}
Computing IFs requires Inverse-Hessian-Vector Products (IHVPs), which are intractable for large networks. Arnoldi-Based IF~\cite{schioppa2022scalingif} projects the Hessian onto a low-dimensional Krylov subspace~\cite{saad2011numerical}:
\begin{equation}\label{eq:arnoldi_based_if}
    H_{\hat{\theta}}^{-1} v \approx G^{\top} \widetilde{H}^{-1} G v = \sum_{i=1}^k\tfrac{1}{\lambda_i}g_i(g_i^{\top}v), 
\end{equation}
where $G$ is a projection, $\widetilde{H}=G H_{\hat{\theta}} G^{\top}$ is projected Hessian induced by Arnoldi process~\cite{saad2011numerical} with $k$ eigenpairs $(\lambda_i,y_i)$, and $g_i=G^{\top}y_i$. This yields efficient Ritz approximations~\cite{saad2011numerical} for IHVPs.

\section{Targeted fine-tuning}
\label{sec:proposed_fine_tuning}
Our methodology combines influence analysis with fine-tuning to improve a deep learning-based receiver performance in four steps, illustrated in Fig.~\ref{fig:high_level} and described below.

As a baseline, we first train the DeepRx model~\cite{honkala2021deeprx} on received frequency-domain waveforms with transmitted bits as labels. Model performance is measured in terms of uncoded Bit Error Rate (BER):
\begin{equation}\label{eq:uncoded_ber}
    \text{BER} = \frac{1}{N_b} \sum_{i=1}^{N_b} \mathbf{1}\{\hat{b}_i \neq b_i\},
\end{equation}
where $N_b$ is the number of transmitted bits, $b_i$ the true bits, $\hat{b}_i$ their estimates by DeepRx, and $\mathbf{1}\{\cdot\}$ is the indicator function.

\begin{figure}[!t]
\centering
\includegraphics[trim={6 8 4 8},clip,width=\columnwidth]{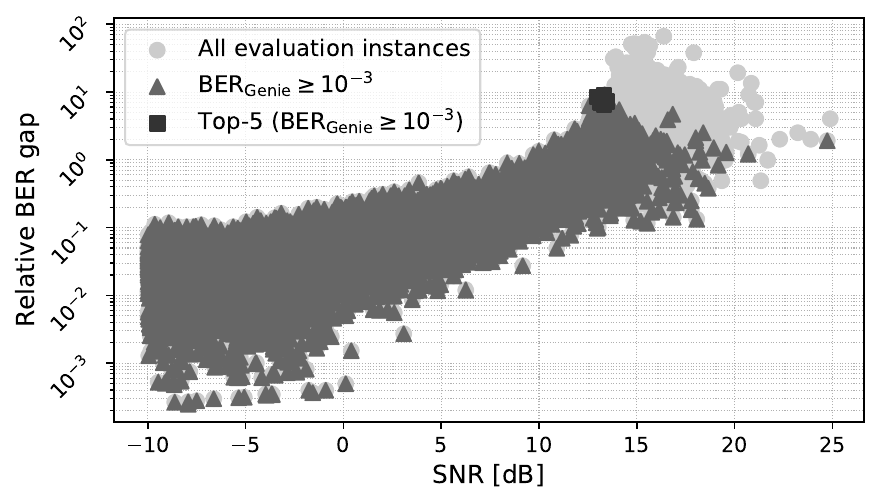}
\caption{Relative BER gaps $\Delta_{\text{BER}}$ as defined in Eq.~\eqref{eq:relative_ber_gap} between DeepRx and a genie-aided LMMSE (full CSI). Targets are the top-5 largest-gap instances in the regime where $\text{BER}_{\text{Genie}} \geq 10^{-3}$.}
\label{fig:selection_of_evaluation_instances}
\end{figure}

\textbf{Step 1: Selecting Target Instances.} We identify evaluation instances whose performance we aim to improve. For each instance, we compute the relative BER gap between the DeepRx output and a genie-aided receiver benchmark:
\begin{equation}\label{eq:relative_ber_gap}
    \Delta_{\text{BER}} = \frac{\text{BER}_{\text{DeepRx}} - \text{BER}_{\text{Genie}}}{\text{BER}_{\text{Genie}}}.
\end{equation}
Instances with the highest $\Delta_{\text{BER}}$ in a practical BER range are chosen as target instances, as illustrated in Fig.~\ref{fig:selection_of_evaluation_instances}.

\textbf{Step 2: Identifying Influential Training Data.} For each target, we compute $\ell$-relative influence scores with Eq.~\eqref{eq:ell_relative_if} using the capacity-like binary cross-entropy loss~\cite[Eq.~9]{korpi2021} (also used in training). The most beneficial training samples--those with the smallest scores--are then selected.

\textbf{Step 3: Fine-Tuning the Receiver.} We adopt influence-guided fine-tuning~\cite{schioppa2023whatifdo}: the receiver is briefly trained on beneficial samples, providing \emph{a quick nudge in the right direction} that reduces the target performance gap without full retraining.

\textbf{Step 4: Evaluation.} Finally, we re-evaluate the fine-tuned receiver. The key metric is the reduction of the BER gap relative to the genie-aided benchmark:
\begin{equation}\label{eq:ber_gap_reduction}
    R_{\text{gap}} = \frac{\Delta_{\text{BER, before}} - \Delta_{\text{BER, after}}}{\Delta_{\text{BER, before}}} \times 100\%,
\end{equation}
where $\Delta_{\text{BER, before}}$ and $\Delta_{\text{BER, after}}$ are the relative BER gaps defined in Eq.~\eqref{eq:relative_ber_gap} before and after fine-tuning, respectively.

\section{Experiments and Results}
\label{sec:experiments}

\subsection{Experimental Setup}
\label{ssec:experimentalsetup}
We re-implemented DeepRx~\cite[Table I]{honkala2021deeprx} in PyTorch~\cite{PytorchNeurIPS2019}, following the original training parameters~\cite[Sec.~V]{honkala2021deeprx} and using the LAMB optimizer with capacity-like Binary Cross-Entropy (BCE) loss from~\cite[Eq.~9]{korpi2021}. Link-level data were generated with Sionna~\cite{hoydis2023sionna} using the parameters of~\cite[Table~1]{tuononen2025nap}, yielding 799{,}968 training and 41{,}112 evaluation samples. Models were trained on both the full dataset (800k) and a reduced set (200k) to study low- and high-training-data regimes. For influence computation, we implemented Arnoldi-based IF, defined in Eq.~\eqref{eq:arnoldi_based_if}, with PyTorch’s JVP. We kept the top-40 eigenvalues, beyond which accuracy gains were negligible
\footnote{\label{additionalresultsnote}Additional material can be found in the Appendix, see Section~\ref{sec:appendix}.}.
Arnoldi used a Krylov subspace of 200, 545 iterations, and batch size 22, resulting in $\sim$2.4M Hessian–vector products (about three passes over the training set), which we found sufficient for stable Hessian approximations.

Target instances were chosen as those with the highest relative BER gap $\Delta_{\text{BER}}$ defined in Eq.~\eqref{eq:relative_ber_gap} among cases with $\text{BER}_{\text{Genie}} \geq 10^{-3}$, since such levels can be already challenging for practical channel decoding. We focused on the top-5 gaps to ensure practical relevance (Fig.~\ref{fig:selection_of_evaluation_instances}). We compared influence computed with BER- and BCE-oriented losses, using both global (classic, Newfluence) and local ($\theta$-relative, $\ell$-relative) variants (Sec.~\ref{sec:background}), and evaluated their effect on BER gap reduction. For BER, we used the smooth surrogate of uncoded BER from~\cite[Eq.~78]{penttinen2025} with cross-loss IF in Eq.~\eqref{eq:cross_loss_if}. For BCE, we used the same capacity-like BCE loss~\cite[Eq.~9]{korpi2021}.

\begin{figure}[!t]
\centering
\includegraphics[trim={4 8 2 4},clip,width=\columnwidth]{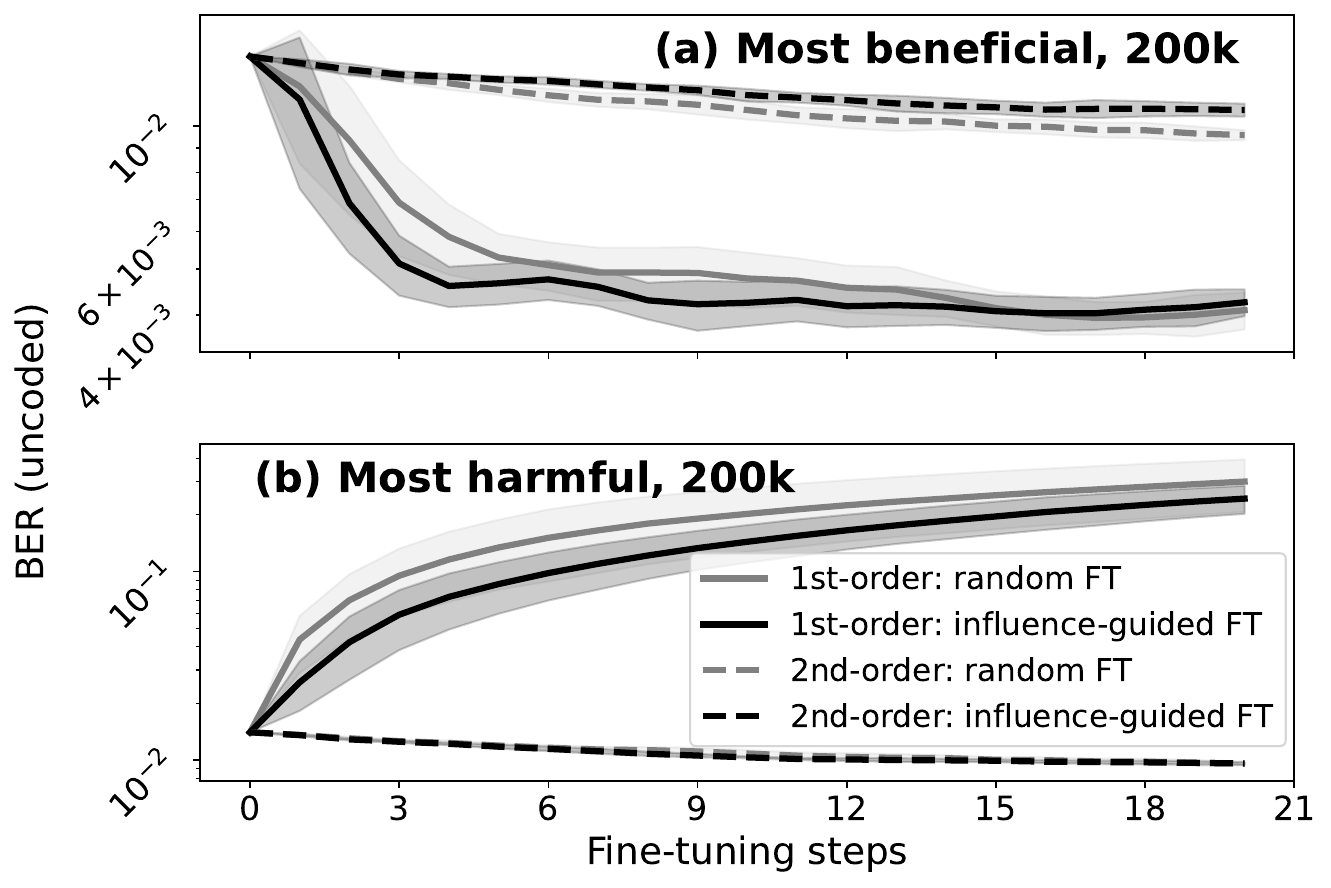}
\caption{BER over fine-tuning steps for first- and second-order methods with random and influence-guided selection: (a) most beneficial and (b) most harmful instances (BCE loss, $\ell$-relative, 10 seeds). Solid lines show mean; shaded areas indicate $\pm 1$ standard deviation.}
\label{fig:fine_tuning}
\end{figure}

Fine-tuning used the same LAMB optimizer with learning rate 0.002 (1/50 of maximum during training), weight decay $10^{-5}$, 24 instances per step, and typically 3 steps per target (1–20 tested)\footref{additionalresultsnote}. We compared random selection vs.\ influence-guided selection from the top-50 most beneficial or harmful instances\footref{additionalresultsnote}. First-order updates were gradient descent for beneficial and naive gradient ascent for harmful cases, since label-correction methods~\cite{li2024deltainfluence} are not applicable to our multi-label classification\footref{additionalresultsnote}. As a second-order alternative, we applied influence-aligned updates using IHVPs from Arnoldi iteration (Eq.~\ref{eq:arnoldi_based_if}), extending gradient descent with Hessian information\footref{additionalresultsnote}. Performance was evaluated via BER gap reduction $R_{\text{gap}}$~(Eq~\ref{eq:ber_gap_reduction}) and BER curves across SNRs for single- and multi-target fine-tuning.

\begin{table}[!t]
\renewcommand{\arraystretch}{1.3}
\caption{Comparison of BER gap reductions $R_{\text{gap}}$ for the highest relative BER gap $\Delta_{\text{BER}}$ target across loss functions, variants, and training sample sizes (1st order, beneficial, 3 steps, 10 seeds). Best results in \textbf{bold}, second-best in \textit{italics}, and selected variant in \textbf{bold}.}
\label{table:ber_gap_reductions}
\centering
\begin{tabular}{c|c|c|c}
    \hline
    \bfseries IF Loss & \bfseries IF Variant (Eq.) & \bfseries 200k samples & \bfseries 800k samples \\
    \hline\hline
    \multirow{4}{*}{\makecell{BER  \\ \cite[Eq. 78]{penttinen2025} \\ using \\ CLIF (Eq.~\ref{eq:cross_loss_if})}} 
      & Classic \eqref{eq:classic_if}                       & 16.3$\;\pm\;$11.5\%           & 25.2$\;\pm\;$14.5\%           \\
    \cline{2-4}
      & Newfluence \eqref{eq:newfluence_if}                 & 50.2$\;\pm\;$10.6\%           & 13.5$\;\pm\;$17.9\%            \\
    \cline{2-4}
      & $\theta$-relative \eqref{eq:theta_relative_if}      & 57.5$\;\pm\;$10.1\%           & \textit{37.4$\;\pm\;$13.9\%}           \\
    \cline{2-4}
      & $\ell$-relative \eqref{eq:ell_relative_if}          & 64.4$\;\pm\;$13.1\%           & 35.2$\;\pm\;$10.1\%           \\
    \hline
    \multirow{4}{*}{\makecell{BCE \\ \cite[Eq. 9]{korpi2021}}} 
      & Classic \eqref{eq:classic_if}                       & \textbf{68.3$\;\pm\;$7.3\%}   & 11.3$\;\pm\;$41.9\%          \\
    \cline{2-4}
      & Newfluence \eqref{eq:newfluence_if}                 & 46.4$\;\pm\;$16.2\%           & 22.2$\;\pm\;$6.9\%            \\
    \cline{2-4}
      & $\theta$-relative \eqref{eq:theta_relative_if}      & 58.9$\;\pm\;$8.3\%            & 10.4$\;\pm\;$13.6\%            \\
    \cline{2-4}
      & \textbf{$\ell$-relative \eqref{eq:ell_relative_if}} & \textit{65.6$\;\pm\;$13.5\%}  & \textbf{47.8$\;\pm\;$8.5\%}     \\
    \hline
\end{tabular}
\end{table}

\begin{figure*}[!t]
\centering
\includegraphics[width=0.98\textwidth]{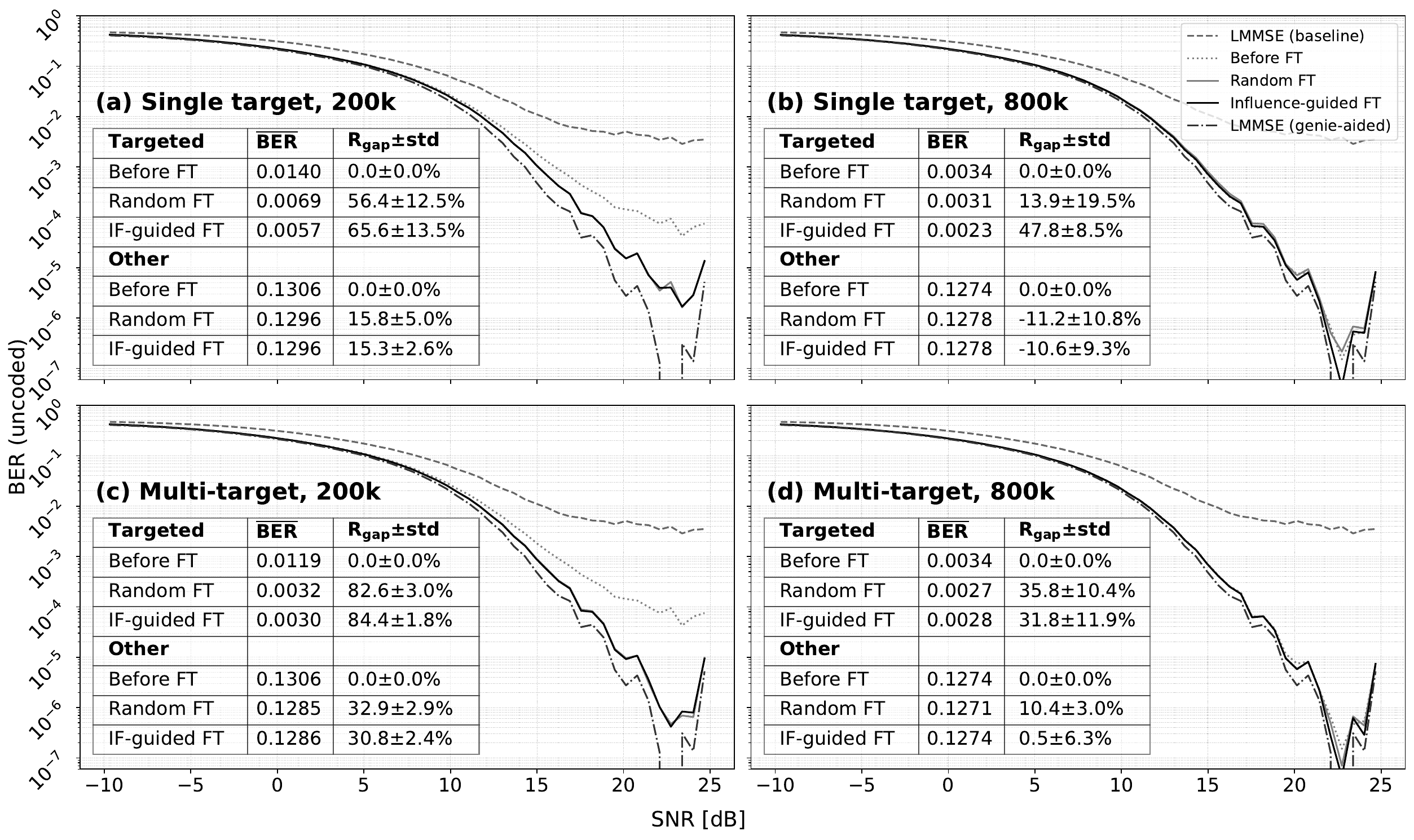}
\caption{Fine-tuning results with first-order method and influence-guided selection of the most beneficial instances. Uncoded BER vs.~SNR (55 bins) is shown for (a) single, 3 steps, 200k; (b) single, 3 steps, 800k; (c) multiple, 15 steps, 200k; and (d) multiple, 15 steps, 800k. Each plot compares baseline LMMSE, before fine-tuning, random fine-tuning, influence-guided fine-tuning, and genie-aided LMMSE. Tables report mean absolute BER and relative BER gap reduction for targeted and other validation instances (BCE, $\ell$-relative; calculated over 10 seeds).}
\label{fig:results}
\end{figure*}

\subsection{Results}
\label{ssec:results}
\textbf{Optimization method and instance selection.} Fig.~\ref{fig:fine_tuning}a and~\ref{fig:fine_tuning}b show that first-order optimization worked well on beneficial instances, reducing BER more quickly than random selection. In contrast, applying it to harmful instances caused catastrophic degradation. Second-order influence-aligned updates were more stable but delivered only negligible gains on both beneficial and harmful cases. Moreover, influence quickly became outdated: after about fifteen fine-tuning steps, the BER of influence-guided and random strategies converged (Fig.~\ref{fig:fine_tuning}a). These results confirm that beneficial influential samples provide a genuine initial advantage, but their effect is short-lived.

\textbf{Comparison across losses and IF variants.} Motivated by the above, we fixed fine-tuning to three steps and compared BER gap reductions across loss functions, IF variants, and training sample sizes (Table~\ref{table:ber_gap_reductions}). Among the tested combinations, BCE with $\ell$-relative IF gave the most consistent reductions for both 200k and 800k samples and was therefore chosen for further experiments.

\textbf{Fine-tuning performance.} With this setup, we evaluated first-order influence-guided fine-tuning on beneficial instances with BCE loss and $\ell$-relative IF (Fig.~\ref{fig:results}). In the single-target setting, it consistently outperformed random tuning on targeted instance (e.g., $47.8\%\pm8.5\%$ vs.\ $13.9\%\pm19.5\%$ with 800k samples). On non-targets, both methods improved BER with 200k but worsened it with 800k, performing similarly overall. The multi-target setting was less favorable: influence guidance only matched random tuning on targets and was consistently worse on non-targets. Thus, it was effective for single-target adaptation but did not generalize to multi-target scenarios. Visual inspection of BER curves further showed benefits at high SNRs and low-BER regimes with 800k samples, where averages understated its effect.

\section{Discussion}
\label{sec:discussion}
Our results demonstrate that influence functions can improve deep learning-based receivers through targeted fine-tuning. By selecting poorly performing evaluation instances and retraining on their most beneficial training samples, we reduced the BER gap relative to a genie-aided benchmark and achieved faster gains than with randomly chosen samples. In this work, selection was restricted to the top-50 beneficial or harmful instances, randomly ordered. This ensured diversity but did not fully exploit the ranking of influence magnitudes. Future work could test whether larger pools or sequential selection provide more stable or stronger improvements.

The choice of loss for influence computation remains an open question. Although a BER-oriented loss is theoretically better aligned with system-level objectives, it did not outperform the training-time BCE loss in our experiments. This points to further exploration of alternative loss metrics and selection criteria, for example capacity-based gap measures instead of relative BER gap.

We were not able to leverage harmful instances effectively: naive gradient ascent destabilized training, and influence-aligned ascent showed no benefit. While machine unlearning methods such as~\cite{li2024deltainfluence} address related challenges, they are not directly applicable to our multi-label setting\footref{additionalresultsnote}. Designing principled strategies to exploit harmful samples remains open. Similarly, second-order influence-aligned updates for beneficial instances proved ineffective, suggesting that better Hessian-based approaches are still needed.

In multi-target settings, influence-guided tuning offered limited benefit, often matching but not surpassing random tuning. Group-based~\cite[Eq.4]{koh2019groupif} and joint~\cite[Eq.3]{hammoudeh2024} approaches may yield more coherent multi-target updates. Beyond this, advanced fine-tuning such as LoRA~\cite{hu2021lora} with PEFT~\cite{peft2022} and trajectory-based influence estimators like TracIn~\cite{pruthi2020tracin} remain promising research directions.

Beyond fine-tuning, influence functions could enhance interpretability by revealing which training subsets or input features (e.g., subcarriers, time slots) most affect predictions~\cite{penttinen2025,bae2022if}--potentially linking patterns to communication-theoretic effects (e.g., frequency selectivity, pilot contamination). They also suggest data-efficient training (pretraining on smaller subsets, refining with influential samples) and on-the-fly adaptation--supporting more robust and efficient deployment in dynamic wireless environments.

\section{Conclusions}
\label{sec:conclusions}
We presented the first application of influence functions to deep learning-based wireless receivers and introduced targeted influence-guided fine-tuning in the wireless domain. On DeepRx, adapting poorly performing instances with their most beneficial training data reduced the BER gap to a genie-aided benchmark more effectively than random fine-tuning, demonstrating practical value.

Open questions remain on (i) the choice of influence loss, (ii) criteria for selecting poorly performing instances, (iii) how to exploit harmful instances in fine-tuning, and (iv) how to handle multiple targets jointly. Promising future directions include lightweight adaptation methods (e.g., LoRA with PEFT), alternative influence estimators (e.g., TracIn), and group- or joint-update strategies.

Influence functions also enhance interpretability by linking predictions to critical training data and suggest new training strategies, such as training on smaller subsets and refining with influential samples, or rapidly adapting to new channel conditions without full retraining. Taken together, they lay the foundation for influence-guided, adaptive, and interpretable neural receivers in 6G.

\bibliographystyle{IEEEbib}
\bibliography{references}

\input{appendix}

\end{document}

%% file: appendix.tex
\section{Appendix}
\label{sec:appendix}

\subsection{Selection of eigenvalues -- Error Analysis}
\label{ssec:eigenvector_selection}
\begin{figure}[p]
\centering
\includegraphics[width=\columnwidth]{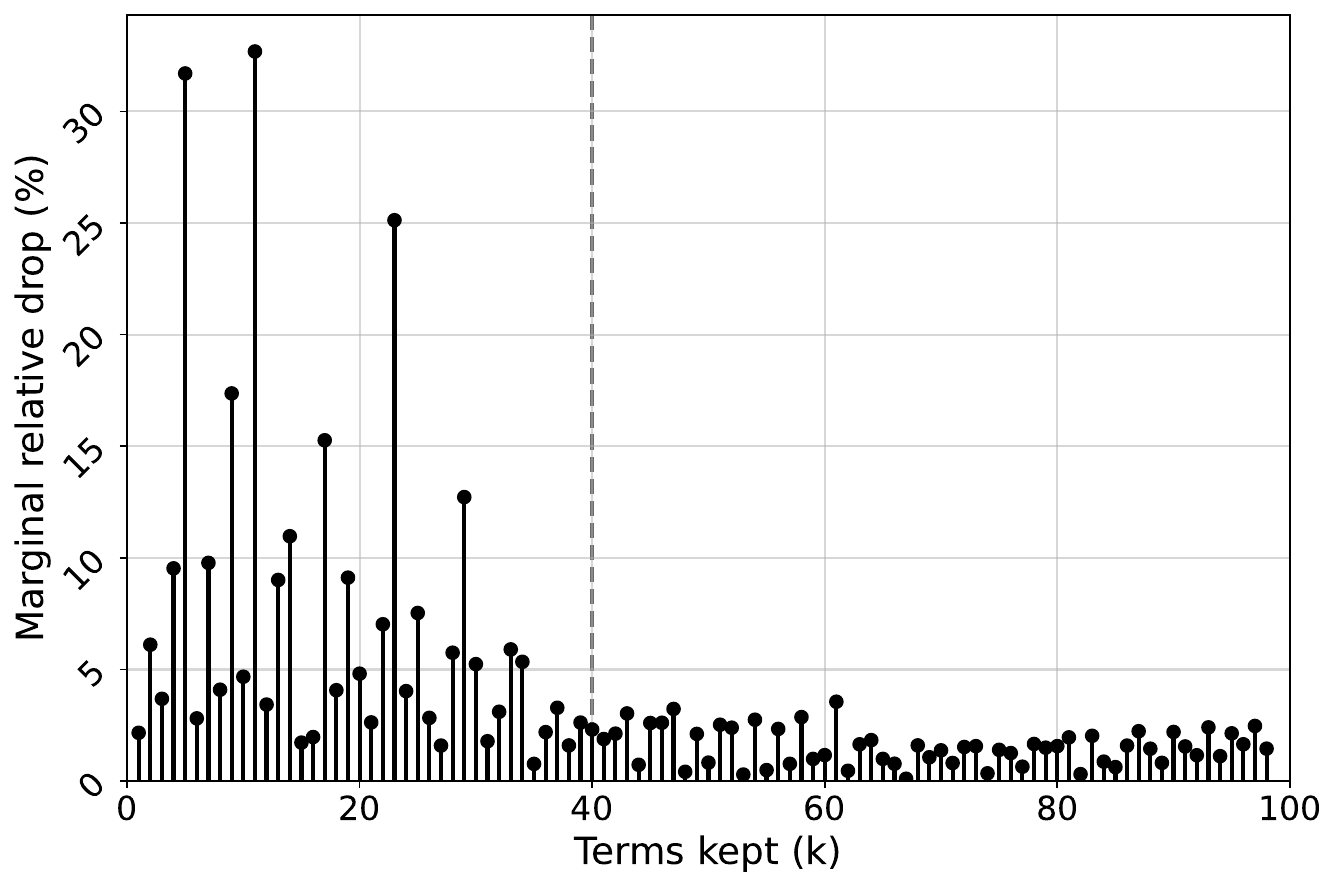}
\caption{Relative reduction in the truncation error upper bound per added term ($k$), computed from \cite[Lemma 1]{schioppa2022scalingif} with $\|\theta\|=1$. The dashed line marks $k=40$, the number of eigenvalues selected.}
\label{fig:selection_of_eigenvectors_relative_error}
\end{figure}

\begin{figure}[p]
\centering
\includegraphics[width=\columnwidth]{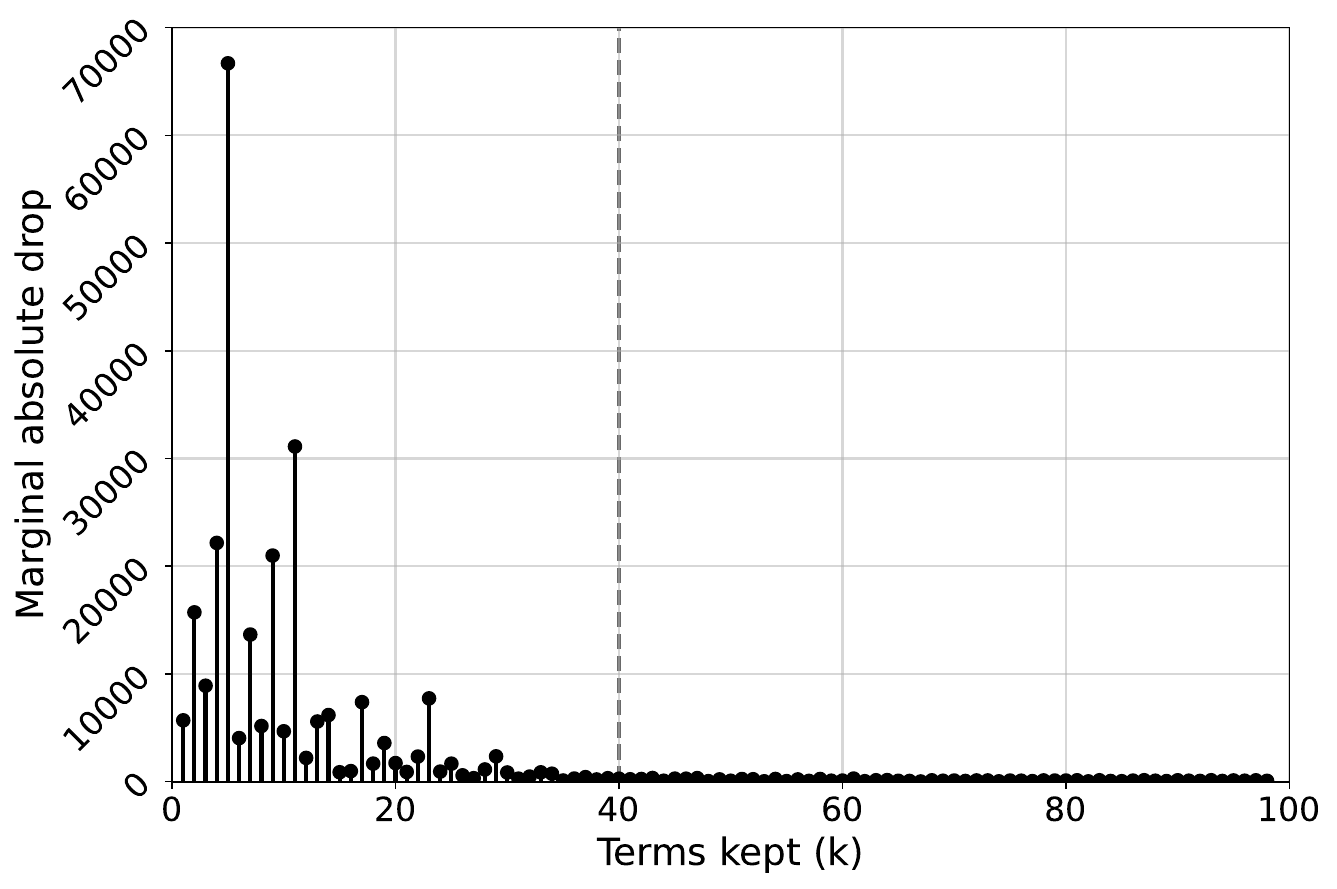}
\caption{Absolute reduction in the truncation error upper bound per added term ($k$), computed from \cite[Lemma 1]{schioppa2022scalingif} with $\|\theta\|=1$. The dashed line marks $k=40$, the number of eigenvalues selected.}
\label{fig:selection_of_eigenvectors_absolute_error}
\end{figure}

For IHVPs we retained the top-40 Arnoldi eigenvalues\footnote{Strictly speaking, Arnoldi returns \emph{Ritz values}: exact eigenvalues of the Hessenberg matrix T, which approximate the eigenvalues of the Hessian H. For simplicity, we refer to them as ‘eigenvalues’ here.}. Using the truncation error bound from~\cite[Lemma 1]{schioppa2022scalingif} (with $\|\theta\|=1$), we measured the marginal reduction per added term. Many of the first 40 terms yielded substantial relative drops (up to $\sim$33\%), whereas beyond 40 terms the relative gains plateau below 4\% (Fig.~\ref{fig:selection_of_eigenvectors_relative_error}). Absolute marginal drops likewise stabilize within the first 40 terms (Figure~\ref{fig:selection_of_eigenvectors_absolute_error}). Based on this elbow, we fix $k=40$ for all experiments.

\begin{figure}[p]
\centering
\includegraphics[width=\columnwidth]{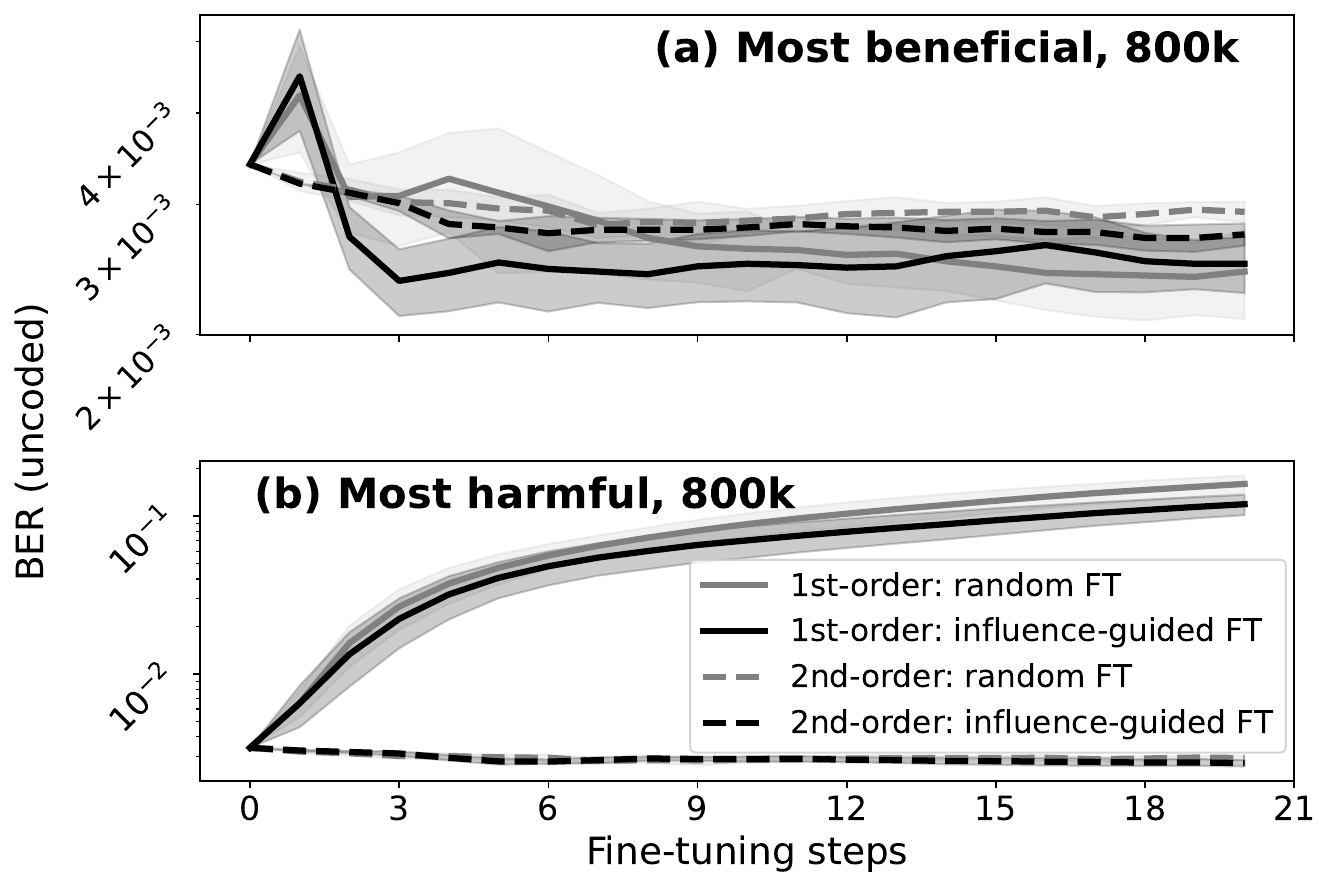}
\caption{BER over fine-tuning steps for first- and second-order methods with random and influence-guided selection: (a) most beneficial and (b) most harmful instances (BCE loss, $\ell$-relative, 10 seeds). Solid lines show mean; shaded areas indicate $\pm 1$ standard deviation.}
\label{fig:fine_tuning_full}
\end{figure}

\begin{figure*}[t]
\centering
\includegraphics[width=\textwidth]{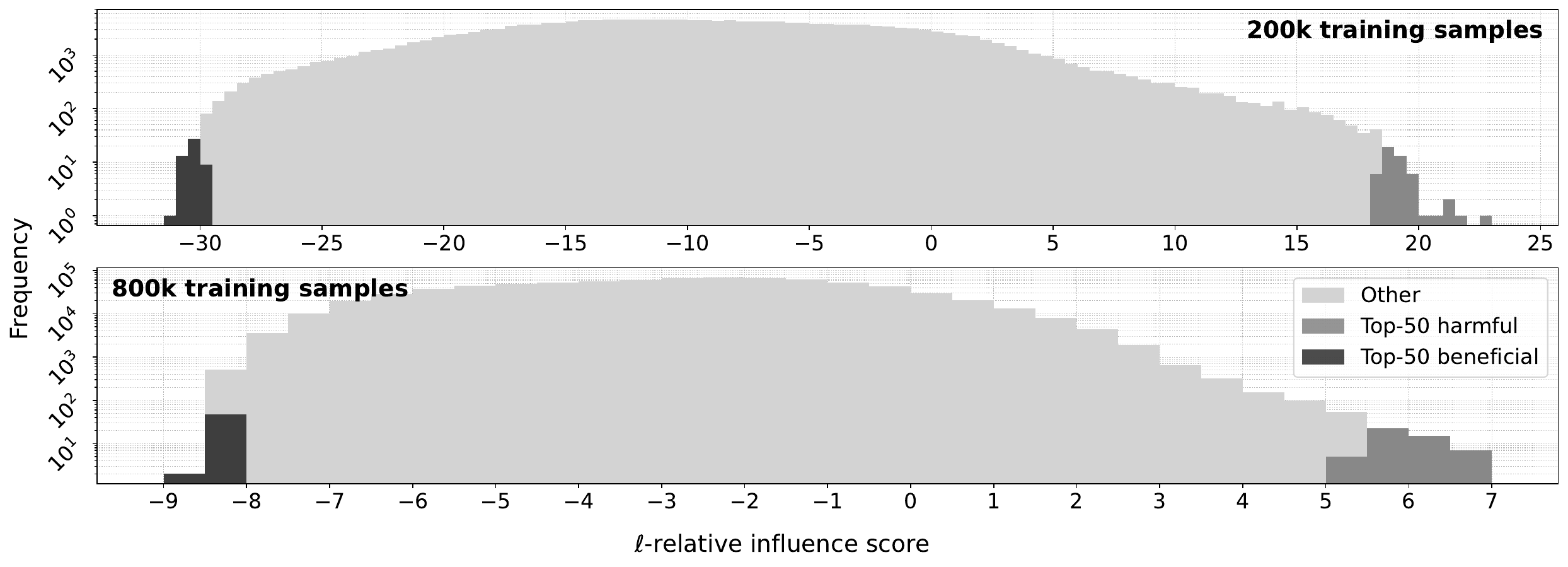}
\caption{Influence score histograms for the worst evaluation instance (highest relative BER gap, $\mathrm{BER}_{\text{Genie}} \geq 10^{-3}$) with (a) 200k and (b) 800k training samples. Light gray: all training samples; medium: 50 most harmful; dark: 50 most beneficial (BCE, $\ell$-relative).}
\label{fig:influence_scores_hist}
\end{figure*}

\begin{figure}[p]
\centering
\includegraphics[width=\columnwidth]{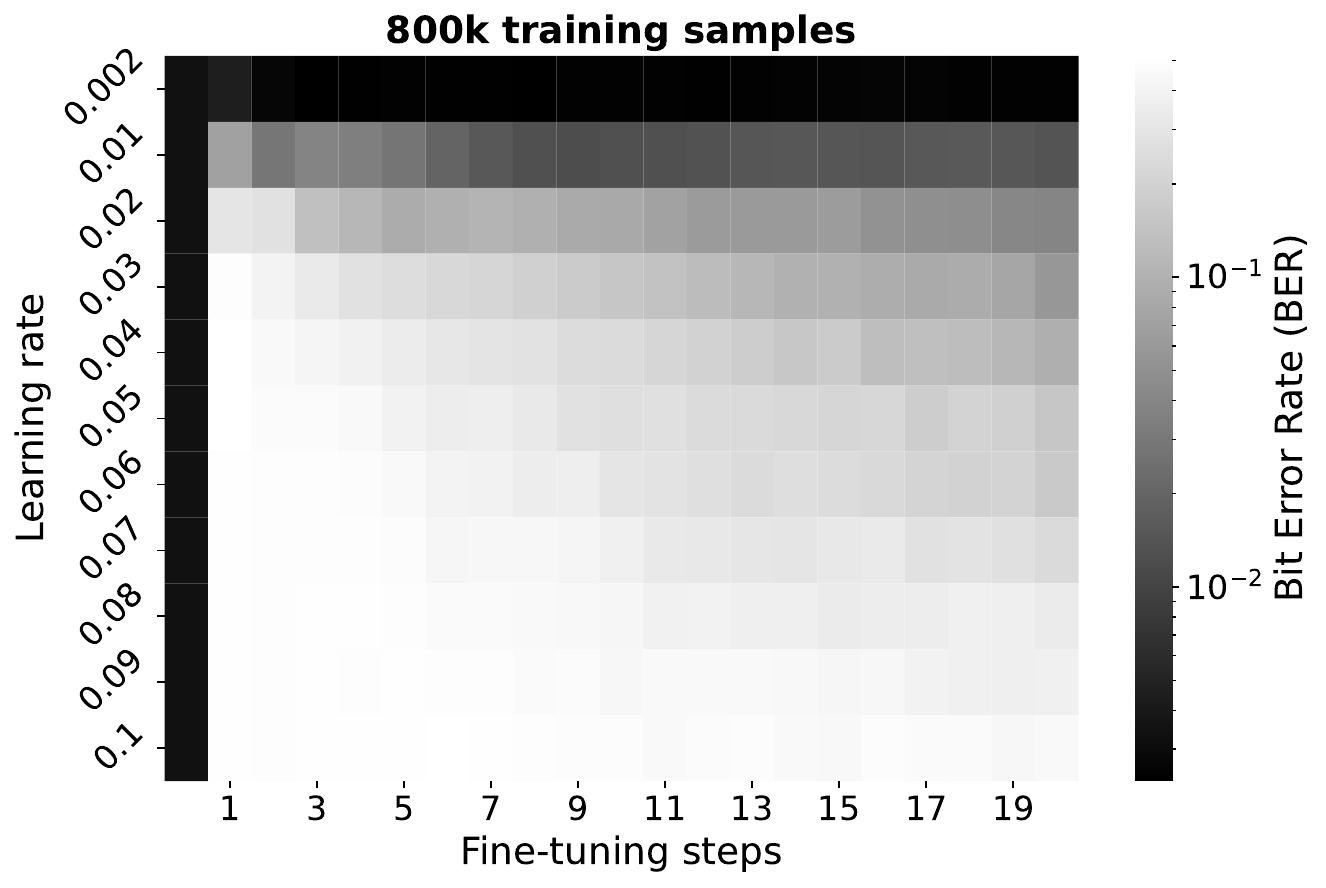}
\caption{Fine-tuning BER for the worst evaluation instance with learning rates 0.01–0.1, (incl. 0.002 used in main experiments). Heatmap over 0–20 steps (BCE, $\ell$-relative, 10 seeds).}
\label{fig:learning_rate_search_coarse}
\end{figure}

\begin{figure}[p]
\centering
\includegraphics[width=\columnwidth]{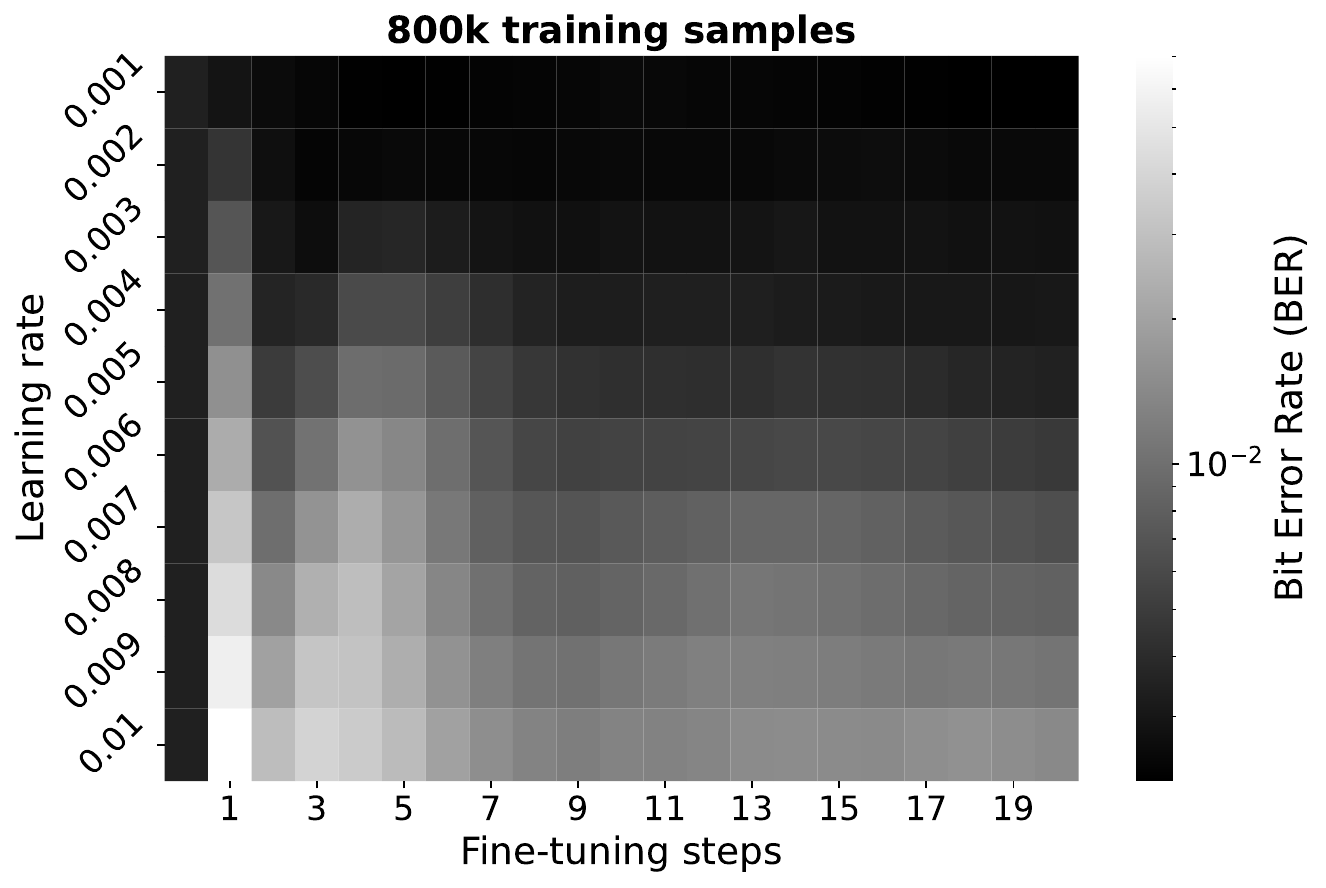}
\caption{Fine-tuning BER for the same instance with learning rates 0.001–0.01. Heatmap over 0–20 steps (BCE, $\ell$-relative, 10 seeds).}
\label{fig:learning_rate_search_fine}
\end{figure}

\subsection{Influence-Aligned (2nd order) Fine-Tuning and Hessian-Based Updates}
\label{ssec:influence_aligned_fine_tuning}
For single-target fine-tuning, we define the influence-aligned second-order parameter update as
\begin{equation}\label{eq:second_order_param_update}
    \Delta \hat{\theta}=\mathrm{sign}(\mathcal{I}_{\text{Classic}}(z_{\text{test}}, z_i))\eta H_{\hat{\theta}}^{-1}\nabla_{\theta}\ell(z_i, \hat{\theta}),
\end{equation}
where $\mathrm{sign}(\cdot)$ is a function returning 1 or -1 based on the sign of the argument, $z_i$ is a training instance influential on test instance $z_{\text{test}}$ and $0<\eta<1$ is the learning rate hyperparameter. The corresponding first-order change (reduction) in test loss is
\begin{align}
    \Delta \ell(z_{\text{test}}, \hat{\theta})
    &\approx \nabla_{\theta}\ell(z_{\text{test}}, \hat{\theta})^{\top}\Delta \hat{\theta} \quad \text{(via Eq.~\ref{eq:second_order_param_update} and Eq.~\ref{eq:classic_if})} \\
    &= -\mathrm{sign}(\mathcal{I}_{\text{Classic}}(z_{\text{test}}, z_i))\eta\mathcal{I}_{\text{Classic}}(z_{\text{test}}, z_i) \\
    &= -\eta |\mathcal{I}_{\text{Classic}}(z_{\text{test}}, z_i)|\\
    &< 0.
\end{align}
Thus, the first update theoretically reduces the loss for $z_{\text{test}}$. This result holds only for the first step and for a single influential training instance, since the Hessian (and hence the IHVPs) change as parameters update. In practice, we reuse the IHVPs already computed for influence estimation, assuming they remain approximately valid for the first few fine-tuning steps. This reuse avoids any extra computation. Updates from multiple influential instances are aggregated within a minibatch and applied jointly.

For multi-target fine-tuning, the update becomes
\begin{equation}\label{appeq:multi_param_update}
    \Delta \hat{\theta}=\eta \sum_{(j, k)\in M}\mathrm{sign}(\mathcal{I}_{\text{Classic}}(z_{\text{j}}, z_k))H_{\hat{\theta}}^{-1}\nabla_{\theta}\ell(z_k, \hat{\theta}),
\end{equation}
where $J$ is the set of targeted test instances, $K$ their selected influential training instances, and 
$M \subseteq J \times K$ the chosen test–train pairs. The resulting first-order change in total targeted loss is
\begin{align}
    \Delta \Bigg(\sum_{j\in J}\ell(z_j)\Bigg) &\approx \sum_{j\in J}\Bigg(\nabla_{\theta}\ell(z_{\text{j}}, \hat{\theta})^{\top}\Bigg)\Delta \hat{\theta} \\
    &= -\eta \sum_{(j, k)\in M} |\mathcal{I}_{\text{Classic}}(z_{\text{j}}, z_k)| \\
    &< 0,
\end{align}
under the simplifying assumption that each influential training instance affects only its paired test instance. Although derived here for the classic influence variant (Eq.~\ref{eq:classic_if}), analogous forms hold for the other IF variants with appropriate scaling applied.

\subsection{Multi-Label Classification and Label-Correction Limitations}
\label{ssec:label_correction_limitations}
DeepRx~\cite{honkala2021deeprx} predicts thousands of binary outcomes (bits) in parallel. In machine learning terms, this is \emph{multi-label classification}, where each bit is treated as an independent binary prediction~\cite{tarekegn2024mll}. This differs from:
\begin{itemize}
    \item \textbf{Binary classification:} predicting a single bit.
    \item \textbf{Multi-class classification:} predicting one label from several classes (not applicable here).
\end{itemize}

Unlike standard multi-label classification, DeepRx labels are long sequences of thousands of bits. In label-correction methods such as  DeltaInfluence~\cite{li2024deltainfluence} or~\cite[Sec. 6]{schioppa2023whatifdo}, a harmful instance can be "flipped" from one class to another. In DeepRx, by contrast, flipping would mean selecting which subset of bits to change--an ill-posed task at this scale. Since these sequences come from coded transmissions~\cite{goldsmith2005,tse2005}, arbitrary changes would also violate system consistency, even if DeepRx does not exploit coding explicitly. Hence, harmful-instance handling cannot rely on existing methods of label-correction, motivating our gradient-ascent (Sec.\ref{ssec:experimentalsetup}) and influence-aligned ascent (Sec.\ref{ssec:influence_aligned_fine_tuning}) approaches.

\subsection{Fine-Tuning Hyperparameter Selection}
\label{ssec:fine_tuning_hyperparameters}
We analyze fine-tuning on the worst-performing evaluation instance, selected using the relative BER gap criterion (Eq.~\ref{eq:relative_ber_gap}) within the practical range $\mathrm{BER}_{\text{Genie}} \geq 10^{-3}$. Figure~\ref{fig:influence_scores_hist} shows the corresponding influence score distributions for both the reduced (200k) and full (800k) training sets. The distributions are highly skewed, with only a few samples exerting substantial influence. Therefore, we focus on the 50 most beneficial and 50 most harmful training instances, following prior work \cite{schioppa2023whatifdo,penttinen2025}.

To study the effect of the learning rate during fine-tuning, we evaluate the same target instance across 0–20 steps and varying rates. Figure~\ref{fig:learning_rate_search_coarse} considers rates between 0.01 and 0.1, including 0.1 (maximum used in original training) and 0.002 (used in our main experiments). Figure~\ref{fig:learning_rate_search_fine} examines the lower range 0.001–0.01. The heatmaps indicate that small learning rates (0.001–0.002) are most effective when fine-tuning for only a few steps.